\documentclass{article}

\usepackage{arxiv}

\usepackage[utf8]{inputenc} % allow utf-8 input
\usepackage[T1]{fontenc}    % use 8-bit T1 fonts
\usepackage{hyperref}       % hyperlinks
\usepackage{url}            % simple URL typesetting
\usepackage{booktabs}       % professional-quality tables
\usepackage{amsfonts}       % blackboard math symbols
\usepackage{nicefrac}       % compact symbols for 1/2, etc.
\usepackage{microtype}      % microtypography
\usepackage{lipsum}		% Can be removed after putting your text content
\usepackage{graphicx}
\usepackage{natbib}
\usepackage{doi}

\usepackage{amsmath}
\usepackage{amssymb}
\usepackage{algorithm}
\usepackage{algpseudocode}
\usepackage{tikz}
\usepackage{graphicx,subfigure}
\usepackage{makecell}
\usetikzlibrary{shapes.arrows}
\usetikzlibrary{calc}
\usepackage{array,multirow}

\usepackage{multicol}

\title{Extreme Multilabel Classification for Specialist Doctor Recommendation with Implicit Feedback and Limited Patient Metadata}

\date{} 					% Or removing it

\author{ \href{https://orcid.org/0000-0003-1290-4798}               {\includegraphics[scale=0.06]{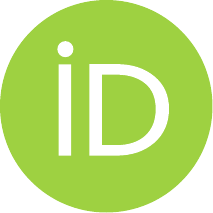}\hspace{1mm}Filipa Valdeira} \\
	Department of Computer Sceince\\
	NOVA School of Science and Technology\\
	Caparica, Portugal \\
	\texttt{filipa.marreiros@unimi.it} \\
	\And
    \href{https://orcid.org/0000-0002-5656-9189}{\includegraphics[scale=0.06]{orcid.pdf}\hspace{1mm}Stevo Racković} \\
	Department of Mathematics\\
	Instituto Superior Técnico\\
	Lisbon, Portugal \\
	\And
    \hspace{1mm}Valeria Danalachi \\
	Department of Computer Sceince\\
	NOVA School of Science and Technology\\
	Caparica, Portugal \\
    \And
	\href{https://orcid.org/0000-0002-6044-4530}{\includegraphics[scale=0.06]{orcid.pdf}\hspace{1mm}Qiwei Han} \\
	Data Science Knowledge Center\\
	NOVA School of Business and Economics\\
	Lisbon, Portugal \\
	\And
	\href{https://orcid.org/0000-0003-3071-6627}{\includegraphics[scale=0.06]{orcid.pdf}\hspace{1mm}Cláudia Soares} \\
	Department of Computer Sceince\\
	NOVA School of Science and Technology\\
	Caparica, Portugal \\	
}

\renewcommand{\shorttitle}{XML Classification for Specialist Doctor Recommendation}

\hypersetup{
pdftitle={Extreme Multilabel Classification for Specialist Doctor Recommendation with Implicit Feedback and Limited Patient Metadata},
pdfauthor={Filipa Valdeira, Stevo Racković, Valeria Danalachi, Qiwei Han, Cláudia Soares}
}

\begin{document}
\maketitle

\begin{abstract}
Recommendation Systems (RS) are often used to address the issue of medical doctor referrals. However, these systems require access to patient feedback and medical records, which may not always be available in real-world scenarios. Our research focuses on medical referrals and aims to predict recommendations in different specialties of physicians for both new patients and those with a consultation history. We use Extreme Multilabel Classification (XML), commonly employed in text-based classification tasks, to encode available features and explore different scenarios. While its potential for recommendation tasks has often been suggested, this has not been thoroughly explored in the literature. Motivated by the doctor referral case, we show how to recast a traditional recommender setting into a multilabel classification problem that current XML methods can solve. Further, we propose a unified model leveraging patient history across different specialties. Compared to state-of-the-art RS using the same features, our approach consistently improves standard recommendation metrics up to approximately $10\%$ for patients with a previous consultation history. For new patients, XML proves better at exploiting available features, outperforming the benchmark in favorable scenarios, with particular emphasis on recall metrics. Thus, our approach brings us one step closer to creating more effective and personalized doctor referral systems. Additionally, it highlights XML as a promising alternative to current hybrid or content-based RS, while identifying key aspects to take into account when using XML for recommendation tasks.
\end{abstract}

% ####################################################################################################################

\section{Introduction}

% Context and importance of the problem
Selecting a suitable doctor for patients significantly affects patient health outcomes~\cite{article:HealthDoc_CF_Qiwei}. During the referral process, a primary care physician (PC) should recommend an appropriate specialist physician according to the individual needs of the patient. However, a PC may struggle to meet this requirement due to limited consultation time, partial knowledge of all matching physicians, infrequent previous patient contact, and potential bias from the doctor's social network~\cite{article:HealthDoc_Regina}. Therefore, an additional tool is needed to facilitate the patient-centered recommendation of specialist physicians. In this paper we recast a traditional recommender setting into a multilabel classification problem that can be solved by current extreme classification methods. Also, we propose a unified model leveraging patient history across different specialties. %% LAST 2 sentences added by CS to reinforce the novelty

% Recommender systems
Recommender systems (RS) have been widely employed to facilitate these decision-making processes \cite{article:guo2016doctor,article:han2018hybrid, etemadi2022systematic}. These systems fall into three categories: Collaborative Filtering (CF), Content-based (CB), and hybrid approaches. CF assumes that users who have rated items similarly in the past will continue to do so in the future. However, CF requires a large amount of data and suffers from the ``cold-start" problem --- a challenge in RS where the system struggles to make accurate recommendations for new users due to a lack of historical interactions~\cite{bobadilla2013recommender}. This is particularly critical in healthcare, as inaccurate recommendations could harm patient care~\cite{etemadi2022systematic}. On the other hand, CB systems recommend items similar to those the user has preferred in the past without relying on information about other users, yet, these systems often struggle to expand users' interests~\cite{article:HealthDoc_RSContent}. Hybrid approaches combine the strategies used in both CB and CF to take advantage of their different strengths.

% the particular setting of healthcare recommendation and challenges
In the context of medical expert recommendations, specific requirements differentiate this field from other domains that employ recommender systems. Patients usually interact with a significantly smaller set of possible physicians, unlike traditional settings with a large pool of interactions from which to learn. Therefore, while CF methods perform well in other domains, most healthcare methods are CB~\cite{article:HealthDoc_RSContent} or hybrid~\cite{article:HealthDoc_RSMF,article:HealthDoc_CF_Deng}. Furthermore, explainability is vital in healthcare applications since doctors and patients are less likely to trust black-box recommendations. This extends beyond user trust; understanding the factors contributing to a recommendation could potentially inform clinical decision-making~\cite{yang2022explainable}. Furthermore, the limited availability of patient metadata due to privacy regulations presents unique challenges in developing effective recommender systems in this domain~\cite{demotes2019new}.

% our sepcific constraints
In this paper, we address the problem of developing a recommender system for specialist doctor referrals in the specific context of a European private healthcare provider. The system is designed to tackle the cold-start problem and handle limited patient metadata while respecting privacy concerns inherent in healthcare. Our data consists of a database of patient-doctor consultations without explicit patient feedback or access to medical records. Furthermore, we focus on providing recommendations to new patients who are likely unfamiliar with the primary care physician and other physicians in the network. Most existing proposals assume the availability of extensive contextual information on the patient's health condition or explicit feedback, which sets a favorable ground for hybrid or CB approaches. However, our specific scenario deviates substantially from these conditions. Moreover, the cold-start problem poses a significant challenge, even for hybrid RS, and remains relatively unexplored in the literature~\cite{etemadi2022systematic}.

% our proposal XML why
Given these constraints, we explore the potential of Extreme Multilabel Classification (XML) methods~\cite{article:PDSparse,article:FastXML,article:SLEEC} for the recommendation task. XML is an instance of the traditional Multilabel Classification (ML) problem, where the label space, instance, and feature space are extremely large. This setting is beneficial in situations where, for each instance (patient), we wish to predict a subset of the possible labels (specialist physicians) relevant to that training point. Due to the large number of available doctors in the network, the recommendation problem could not be framed in the traditional Multilabel Classification setting, but the emergence of XML alternatives brings this as a possibility.

XML methods have been successfully applied in document classification tasks~\cite{liu2017deep}. Additionally,~\cite{article:XML_review} have suggested that XML offers potential tools for addressing recommendation tasks. 
One of the primary advantages of using XML is that it solves the cold-start problem when a system needs to make predictions for new users. Moreover, XML is beneficial when historical interactions are limited because it centralizes user features. However, there are few applications of XML for recommendation tasks and comparisons with traditional RS in the literature.
Recent proposals have suggested using label metadata (item features), as noted in~\cite{article:XML_DECAF,article:XML_ECLARE}.
Label metadata can be a powerful source of information when user features alone may not be informative.
Our research builds on these proposals and presents a way to approach the doctor recommendation problem as a multilabel classification by considering patient and doctor features. We empirically demonstrate that the feature-centered nature of XML can overcome the limitations of traditional RS, and we test this by comparing our method with state-of-the-art (SOTA) methods.

Therefore, our contributions can be summarized as follows:
\begin{itemize}
    \item We propose a \textbf{solution for the problem of specialist doctor recommendations from limited patient metadata} while respecting privacy concerns inherent to the healthcare context. Our method leverages implicit data, which is easy to retrieve and can handle new users with only basic demographic information;
    \item To formulate the recommendation problem as an XML instance, we address two key challenges: \textbf{encoding existing features in a TF-IDF-like manner} and \textbf{converting consultation history into appropriate labels};
    \item 
    We compare our approach with SOTA recommender systems and show that XML provides \textbf{better predictions for existing users and uses existing features better when predicting for users without previous interactions}. This improved prediction performance demonstrates that XML is suitable for the doctor referral process and highlights its potential as an alternative to traditional RS, which is still understudied in the literature.
\end{itemize}

% ####################################################################################################################

\section{Related work}

\paragraph{Healthcare Recommendation Systems (HRS)}

Doctor recommendation systems, a subset of HRS, are widely explored, covering primary doctors \cite{article:HealthDoc_RSContent,article:HealthDoc_Hybrid_Primary}, specialists \cite{article:HealthDoc_Implicit}, or both \cite{article:HealthDoc_CF_Qiwei,article:HealthDoc_SeekDoc, article:HealthDoc_RSMF,article:HealthDoc_CF_Deng,article:HealthDoc_iDoctor,article:HealthDocHopsit_Narducci}.They vary in focus, with some aiming to predict the best doctors, in general,~\cite{article:HealthDoc_DrRight}, while others, like our work, strive to propose recommendations that meet individual patient needs.
Predicting specialist doctors introduces an extra layer of complexity. Some research circumvents this by creating a separate model for each department~\cite{article:HealthDoc_Implicit}, which leads to data scarcity issues for less visited specialties. In our study, we build a {\bf unified model} that leverages the patient history across different departments.

\paragraph{Data Sources}

The primary data sources considered in HRS are feedback and doctor and patient metadata.

\textit{Feedback.}
Explicit feedback comes from text reviews \cite{article:HealthDoc_RSMF,article:HealthDoc_iDoctor} or ratings \cite{article:HealthDoc_CF_Deng,article:HealthDoc_DrRight} and requires the patients to take on an active role, whereas implicit feedback is derived from consultation history \cite{article:HealthDoc_Regina,article:HealthDoc_CF_Qiwei,article:HealthDoc_SeekDoc,article:HealthDoc_Implicit}. The latter, which forms the basis of our study, is easier to obtain but less informative.
When converting consultation history into feedback, different assumptions can be made. While some studies consider each visit as positive feedback \cite{article:HealthDoc_Implicit}, others assign significance based on recency and frequency \cite{article:HealthDoc_RSContent,article:HealthDoc_Hybrid_Primary}. Our approach uses a weighting scheme to assign labels to each training point.

\textit{Doctor and Patient Metadata.}
Patient metadata, including medical history \cite{article:HealthDoc_SeekDoc,article:HealthDocHopsit_Narducci} or current symptoms \cite{article:HealthDoc_Implicit}, can provide a rich source of information for recommendations. However, due to privacy concerns, we focus on doctor metadata and patient demographic information, similar to \cite{article:HealthDoc_RSMF,article:HealthDoc_CF_Deng}.

\paragraph{Recommendation Approaches}

Most HRS rely on hybrid \cite{article:HealthDoc_iDoctor,article:HealthDoc_RSMF,article:HealthDoc_CF_Deng} or content-based RS \cite{article:HealthDoc_RSContent}. However, these methods often struggle with the cold-start problem, i.e., providing recommendations for new patients \cite{article:HealthDoc_Implicit}. The cold-start problem is critical, as new users are typically the ones most in need of recommendations. Unfortunately, most existing methods require patient feedback or symptom data, which is not always readily available. Under the limited feedback, sparse metadata, and the need for cross-specialty predictions and to cater to new users, traditional RS may fall short. Extreme multilabel classification (XML) can potentially harness the available features more effectively. In contrast to most existing methods, XML is not solely reliant on extensive user history or symptom data, which makes it a promising approach for healthcare recommendation systems. In particular, the unique challenges posed by our problem --- the requirement to make predictions across different specialties and for new users, as well as the limited feedback and metadata --- suggest that XML has the potential to leverage the existing features more effectively than a traditional RS.

% ####################################################################################################################

\section{Dataset and Problem Setting}
\label{sec:dataset_prob}

\paragraph{Dataset}\label{sec:dataset}

The dataset that forms the basis of this study is derived from patient-doctor consultations provided by a private European health network, encompassing $P$ patients and $L$ doctors. Each interaction involves a unique patient $p$ and a doctor $l$, with an assigned time and location. In addition, demographic data about patients and doctors is available, along with the educational qualifications and specializations of the doctors~\cite{article:HealthDoc_Regina}. The dataset encompasses $16$ hospitals, with doctors possibly operating across multiple locations and patients attending different hospitals. More details about the dataset can be found in the Supplementary Material. Although the present models do not include it, temporal information was taken into account when creating the train and test splits. To avoid the model from accessing future information, a particular timestamp $t$ was determined as the cut-off point for each patient $p$. As a result, appointments that occurred before $t$ were assigned to the training set, while those that occurred after $t$ were assigned to the test set.

\paragraph{Problem Setting}

Our main goal is to leverage the consultation histories and available metadata for individual patients and doctors to predict a patient's preferred doctors. This objective holds for both patients observed during the training phase and those who were not. In practical terms, this goal translates into predicting a ranked list of physicians $L$, given the features of a single patient. The solution should align with the patient's preferences and consider the specialist category required by the patient at the given time. 

\paragraph{Recommendation Approach}  Given the absence of explicit labels/ratings in the consultation data, we follow state-of-the-art approaches \cite{article:HealthDoc_Implicit,article:HealthDoc_RSContent,article:HealthDoc_Hybrid_Primary} and consider each interaction between a patient and a doctor as positive feedback. In particular, we assume that if a patient has more interactions with one doctor, this reflects a higher preference. However, when selecting a doctor, each patient does not consider all available doctors but all doctors of the specialty in need. Thus, we build the rating matrix $R \in \mathbb{R}^{P \times L}$ with entries
\begin{equation}
R_{pl} = \frac{n_{pl}}{n_{ps}},\label{eq:rat}
\end{equation}
where $n_{pl}$ is the number of visits of patient $p$ to doctor $l$ and $n_{ps}$ is the number of times patient $p$ visited doctors of specialty $s$, where $s$ is the specialty of~doctor $l$. Note that the absence of interactions is considered negative feedback, a usual assumption when handling implicit feedback. Traditional RS learn preferences from matrix $R$ or directly from the interactions, if they accommodate implicit feedback. At prediction time, each user is assigned a ranked list of items, which expresses preference. In the next section, we show how to formulate the problem in the (Extreme) Multilabel Classification setting instead.

\section{Extreme Multilabel Classification for Recommendation}

In this paper, we recast a traditional recommender setting into a multilabel classification problem that current extreme classification methods can solve. Also, we propose a unified model leveraging patient history across different specialties.

Multilabel classification solves the task of assigning each data point with a small subset of the entire label space. To cast a recommendation problem in this setting, we take data points as users and labels as items. Therefore, assigning the most relevant labels to a data point can be understood as predicting the top items of preference for the user. Below we introduce a formal definition of this problem.

\paragraph{Problem Definition.} For each patient $p$, we define a vector of ground truth labels $\textbf{y}_p \in \{-1, +1 \}^L$, where $L$ is the total number of physicians (labels). The $d^{th}$ component of $\textbf{y}_p$ is $+1$ if doctor $d$ is a relevant label for patient $p$ and $-1$ otherwise. We also define feature vectors, $\textbf{x}^p$ for patients and $\textbf{x}^d$ for doctors. A multilabel classifier will learn to predict $\hat{\textbf{y}}_p$ for new users $\tilde{\textbf{x}}^p$.

\paragraph{From Recommendation to Classification.}

Matrix $R$ is the input to a traditional RS dealing with explicit feedback. However, each patient must be assigned a ground truth vector in a classification setting. If we consider any visit to a doctor a positive label, we do not capture information on the frequency of visits to a specific doctor; consequently, there is no information on the patient's preference amongst the subset of visited doctors. Therefore, we opt for converting the ratings $R$ into labels, with a predefined threshold $R_{min}$. We only consider as labels doctors with a higher rating than $R_{min}$, i.e., we build the label matrix $Y \in \mathbb{R}^{P \times L}$ as
\begin{equation}
    Y_{pl} = 
        \begin{cases} 
        1 \; \text{ if } R_{pl} > R_{min} \\
        0 \; \text{ otherwise}.
        \end{cases}
\end{equation}

For patients and doctors, we have separate feature matrices $X_p, X_d\in \mathbb{R}^{P \times N}$ respectively, consisting of feature vectors $\textbf{x}^p,\textbf{x}^d\in \mathbb{R}^N$ defined for each scenario.

Unlike traditional RS, the predictions are only a (ranked) subset of the entire label space, not the complete list of ranked labels. This is suitable for the application at hand, as we want to recommend a small number of physicians to each patient. Thus, for patient $p$, the prediction vector $\hat{y}_p \in \mathbb{R}^L$ will only have $B \ll L$ non-zero elements, containing the scores of $B$ most relevant doctors for patient $p$. We train the model on the entire dataset, i.e., including all possible doctors regardless of their specialty. Unlike training one model for each specialization, this allows us to benefit from a more significant number of interactions and to transfer knowledge between different specialties. However, the classification method will not distinguish between doctor categories at prediction time. Therefore, to predict for specialties $s$, we filter the $B$ predictions and retrieve the ranked subset of doctors for $s$, so it is possible that for some patients, we cannot predict doctors in some specialties. This is a limitation of the current approach, and it is considered when evaluating the results (see Section~\ref{sec:Results}).

\paragraph{Extreme Multilabel Classification Method.}
Given a user feature matrix $X_p$ and a label matrix $Y$, XML methods learn a classifier that predicts the most relevant subset of labels in $L$ for each data point. While earlier approaches were restricted to these inputs \cite{article:PDSparse,article:FastXML,article:SLEEC}, new methods allow for the inclusion of label metadata as matrix $X_d$ \cite{article:XML_DECAF,article:XML_ECLARE}. Label features have been shown to improve predictions and are pertinent to our problem, as the dataset contains information about doctors. Within these methods, we select DECAF as a representative for the XML category due to its high performance and low computational cost. DECAF first learns an embedding for label and point features and then learns One-vs-All classifiers on those embeddings. These embeddings are learned separately, but label and point features are assumed to have the same dimension and live in the same space, which holds for our feature encoding method. A cluster-based shortlister is also used at training and prediction time to reduce computational complexity. 

% ####################################################################################################################

\section{Features}\label{sec:Method}
\begin{figure}
\centerline{\includegraphics[clip, trim=8cm 5cm 7cm 2.5cm, width=0.5\linewidth]{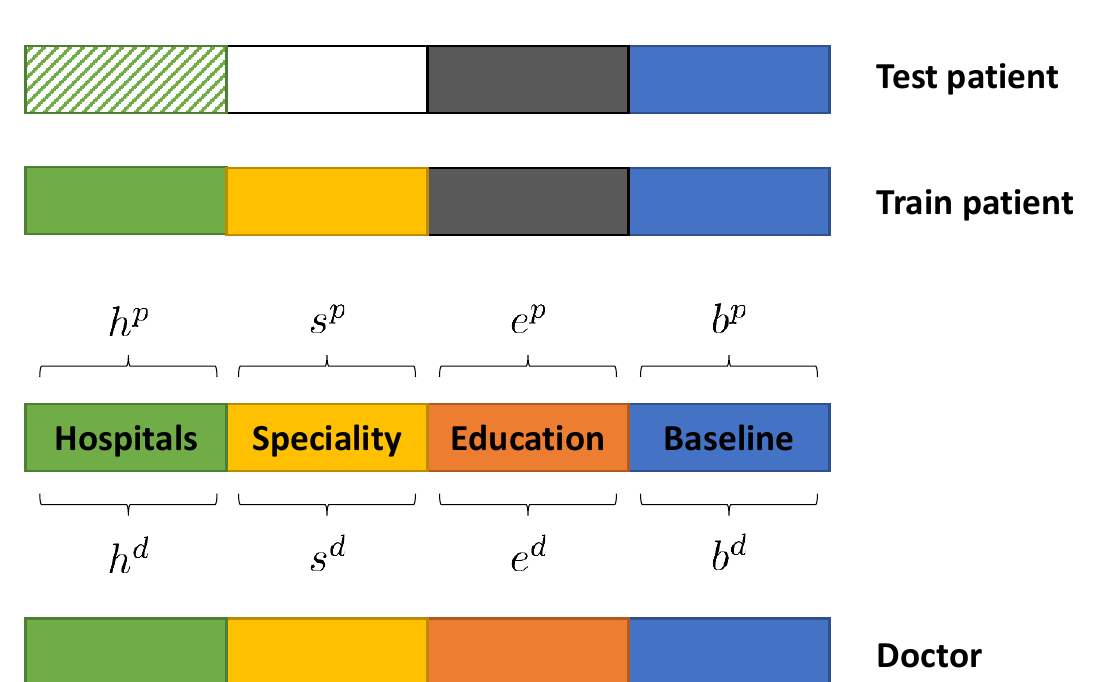}}
\caption{Four distinctive groups of features from consultation data. Empty rectangles indicate that a corresponding feature is not available for that group and it will be masked with a vector of zeros. Grey rectangles are also masked with zeros, but correspond to available information (i.e., patients never have education features). Finally, a hatched green pattern in the case of test patients indicates that this feature is available if calculated as distances but not when the number of visits is considered (see Sec.~\ref{sec:Method_Fatures}).}
\label{fig:features}
\end{figure}

\begin{table}
\centering
\caption{Comparison of the data availability scenarios described in Sec.~\ref{sec:Method_Scenarios}. \textbf{P} stands for patients, and \textbf{D} for doctors.}
\resizebox{0.6\linewidth}{!}{
\begin{tabular}{c|cc|cc|cc|cc|cc}
& \multicolumn{2}{c|}{\textbf{S-1}} & \multicolumn{2}{c|}{\textbf{S-2}} & \multicolumn{2}{c|}{\textbf{S-3}} & \multicolumn{2}{c|}{\textbf{S-4}} & \multicolumn{2}{c}{\textbf{S-5}} \\ \hline
& \textbf{P} & \textbf{D} & \textbf{P} & \textbf{D} & \textbf{P} & \textbf{D} & \textbf{P} & \textbf{D} & \textbf{P} & \textbf{D} \\ \hline
{Baseline} & \checkmark & \checkmark & \checkmark & \checkmark & \checkmark & \checkmark & \checkmark & \checkmark & \checkmark & \checkmark \\ 
{Specialization} &  &  & \checkmark & \checkmark &  &  &  &  & \checkmark & \checkmark \\ 
{Education} &  & \checkmark &  & \checkmark &  & \checkmark &  & \checkmark &  & \checkmark\\ 
{Hospitals (visits)} &  &  &  &  & \checkmark &  \checkmark &  & \checkmark &  & \checkmark \\ 
{Hospitals (distances)} &  &  &  &  &  &  &\checkmark   &  &\checkmark  & \\ 
\end{tabular}
}\label{tab:scenarios}
\end{table}

As noted in the previous section, our approach assumes the existence of feature vectors for both patients and doctors, $\textbf{x}^p$ and $\textbf{x}^d$, respectively. In this section, we show how to convert the information included in the interactions to these feature vectors.

\paragraph{Feature Extraction}\label{sec:Method_Fatures} We derive four distinct feature groups from the interaction data (Fig.~\ref{fig:features}). The first group, the baseline features, is data readily available. The corresponding feature vector is denoted $\textbf{b}^p$ for patients and $\textbf{b}^d$ for doctors. It contains three elements: $b^p_1, b^d_1 \in \{0,1\}$ are binary features that indicate the biological sex of the subject; $b^p_2, b^d_2 \in \mathbb{N}$ are positive integers that indicate the age; and $b^p_3, b^d_3 \in \mathbb{N}$ give the number of occurrences of the patient/doctor in the training set. We use min-max normalization for the last two features, to achieve a similar encoding to TF-IDF.

The second group of features, $\textbf{e}^p, \textbf{e}^d$, pertains to the doctors' education history. For doctors, this categorical data is encoded into binary vectors $\textbf{e}^d\in\{0,1\}^{N_e}$, where $N_e$ is the total number of educational institutions that appear in the training set. The value of the vector is one for those institutions attended by a given doctor and zero otherwise. Since patients have not attended these institutions, $\textbf{e}^p$ will only contain zeros for any patient. However, it is necessary to include it, since patient and doctor features must live in the same feature space.

The third group of features, $\textbf{s}^p, \textbf{s}^d$, encodes the specializations of doctors, as a binary vector $\textbf{s}^d\in\{0,1\}^{N_s}$, where $N_s$ is the total number of specializations in the dataset. These features take on a different meaning for patients, representing the different specializations each patient has attended. Hence, we define $s^p_i$ as the proportion of visits a patient $p$ made to any doctor of the specialization $i$ to the total number of visits the patient made.  Note that this feature will only be present for patients already seen during training (Fig.~\ref{fig:features}), while for new patients, we have $\textbf{s}^p=\textbf{0}.$

The final group features, $h^d, h^p$, refer to the location of appointments. For doctors, the hospital features $\textbf{h}^d$ represent the fraction of their consultations in each hospital. That is, the $i$-th element of $\textbf{h}^d$ is given as the number of times the doctor's interaction occurs in the hospital $i$, normalized in $l_1$. For patients, we consider two possible encodings. \textbf{Encoding 1} follows the approach used for doctors, i.e., $\textbf{h}^d$ reflects the number of visits of each patient to each hospital. Alternatively, we consider \textbf{Encoding 2} that reflects the distance of the patient's address to each hospital. However, while higher values in  $h^d$ reflect a notion of a preferred hospital, distance encoding would convey the opposite meaning. Therefore, we model $\textbf{h}^p$ as a proximity vector, where $h^p_i$ is the inverse of the distance from the patient's location to a hospital $i$ (normalized in $l_{\infty}$). Note that, for the first encoding, only patients already seen in the training will have a non-zero feature vector, while in the latter case, all patients will have an informative feature vector (Fig.~\ref{fig:features}). 

\paragraph{Data Availability Scenarios} \label{sec:Method_Scenarios}  We denote the matrix $X_f \in \mathbb{R}^{P \times N}$ as the patients' features and $D_f \in \mathbb{R}^{L \times N}$ as the doctors' features. We consider five distinct scenarios to assess the impact of different feature groups on the performance of the proposed method (Table \ref{tab:scenarios}). Starting from  baseline and education features, we progressively add specialization and location features as follows: 
\begin{itemize}
  \item \textit{Scenario 1} (S1) includes only the baseline features $\textbf{b}^p, \textbf{b}^d$ and education features $\textbf{e}^p, \textbf{e}^d$. Hence, the patient feature vector for this case is $\textbf{x}^p=[(\textbf{b}^p)^T,(\textbf{e}^p)^T]$, and similar to that of doctors. As we consider these features a minimal representation of the subjects, all the subsequent scenarios will also include them.
  \item \textit{Scenario 2} (S2) consist of features form S1, augmented with the specialty features $\textbf{s}^p, \textbf{s}^d$, i.e., we have $\textbf{x}^p=[(\textbf{b}^p)^T,(\textbf{e}^p)^T,(\textbf{s}^p)^T].$
  \item \textit{Scenario 3} (S3) and \textit{Scenario 4} (S4) augment S1 with hospital proximity features $\textbf{h}^p$ and $\textbf{h}^d$, i.e., we have $\textbf{x}^p=[(\textbf{b}^p)^T,(\textbf{e}^p)^T,(\textbf{h}^p)^T]$. In S3 we use $\textbf{h}^p$ with Encoding 1 (visits) and in S4 we consider $\textbf{h}^p$ with Encoding 2 (distances).
  \item \textit{Scenario 5} is the union of S1, S2 and S4, i.e., $\textbf{x}^p=[(\textbf{b}^p)^T,(\textbf{e}^p)^T,(\textbf{s}^p)^T,(\textbf{h}^p)^T]$. The motivation to select S4 instead of S3 will be given in Section~\ref{sec:Results}.
\end{itemize}
\begin{table}[tb]
\centering
\caption{Dataset characteristics. The first column refers to the training dataset, the second to the seen patients in the test dataset, and the third to new patients in the test set.}\label{tab:datasets}
\begin{tabular}{llll}
\hline
             & Train   & Test seen & Test new  \\ \hline
Interactions & 3914892 & 713730     & 628091       \\
Patients     & 840960  & 324213     & 152498       \\
Doctors      & 1044    & 992        & 999         
\end{tabular}
\end{table}

% ####################################################################################################################

\section{Experimental Results}\label{sec:Results}

In this section, we test the performance of our method for recasting a traditional recommender setting into a multilabel classification problem to be addressed by extreme classification. We show empirically the superiority of our unified model leveraging patient history across different specialties, with respect to SOTA baselines.
We aim to determine which scenario (combination of features) is most suitable and how they impact XML predictions. Additionally, we compare XML with SOTA recommenders, explore the difference between observed and new patients, and analyze performance across different specializations.
\begin{figure}
\centerline{\includegraphics[width=0.4\linewidth]{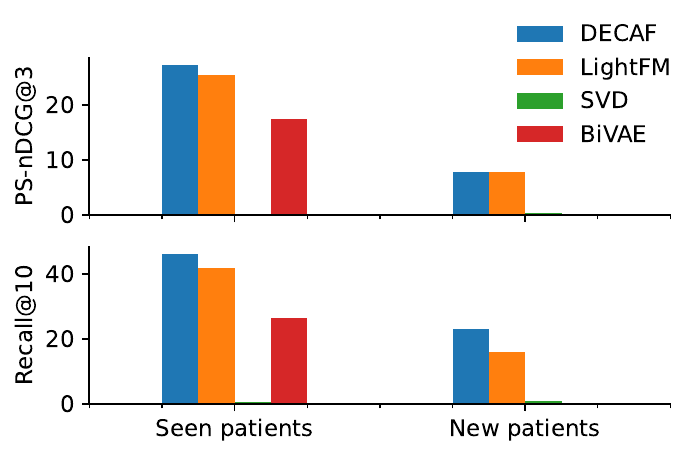}}
\caption{ PS-nDCG@3 and Recall@10 for all methods for seen and new patients. For methods with features (DECAF and LightFM), we select the most favorable scenario per method and case.}
\label{fig:Best_case_all}
\end{figure}

\begin{figure*}[h]
\centerline{\includegraphics[width=0.9\linewidth]{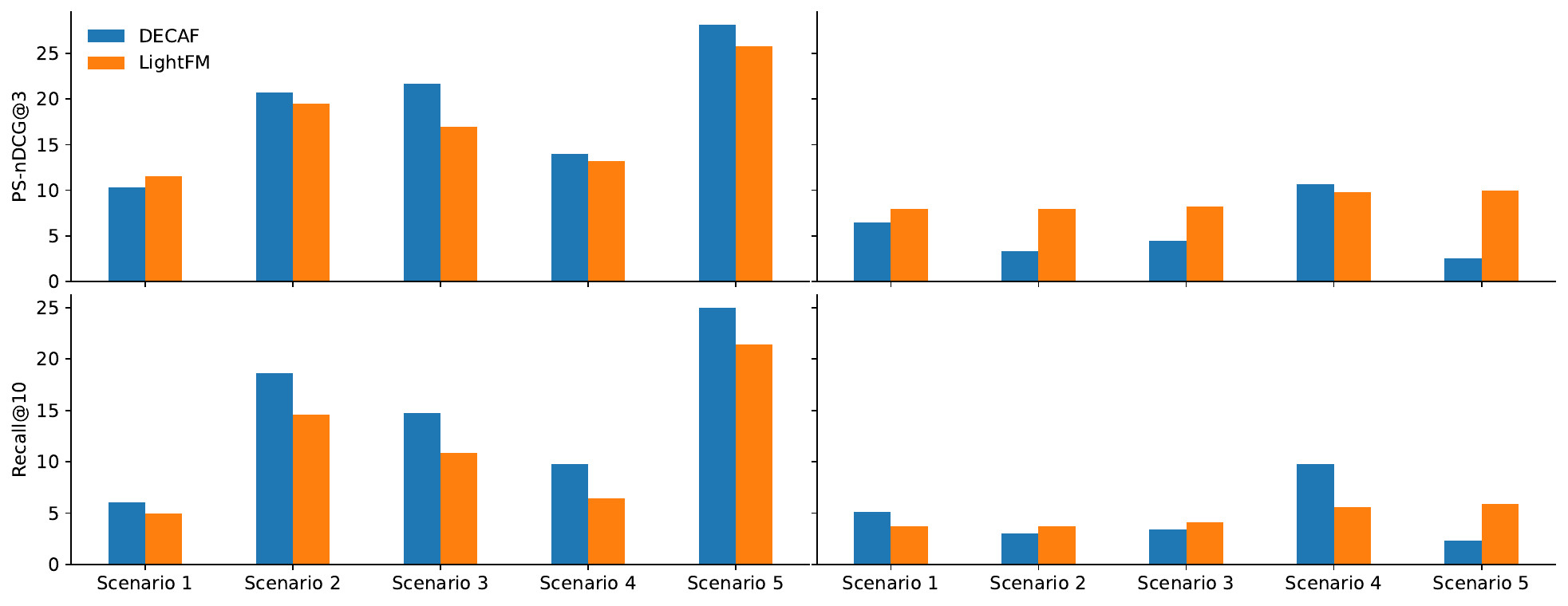}}
\caption{Comparison between DECAF and LightFM for different Scenarios and for seen patients (left) and new patients (right).}
\label{fig:DECAFvslightfm}
\end{figure*}

\subsection{Experimental setting}

\subsubsection{Datasets} We use a real dataset that contains 1064 doctors, 1,003,809 patients, and 2,890,042 interactions. Due to containing personal information, this dataset will not be published. Instead, we will provide simulated data with a compatible distribution. As mentioned in Sec.~\ref{sec:dataset}, the data were divided into train and test sets, considering temporal factors to prevent overfitting. The test set consists of $30\%$ of the data and is divided into two equal parts, covering the following cases. \textbf{Standard RS scenario} where only patients with previous interactions are taken into account, and \textbf{ New patient scenario} where only interactions with previously unseen patients are considered (Table~\ref{tab:datasets}). 

\paragraph{Benchmark methods}

We assess XML against three notable RS models: Singular Value Decomposition (SVD), Bilateral Variational Autoencoder (BiVAE)~\cite{truong2021bilateral}, and LightFM~\cite{kula2015metadata}. Although SVD is a well-established model, BiVAE and LightFM are SOTA models with advanced features, offering a blend of collaborative filtering and hybrid RS capabilities. For collaborative RS, we give as input the rating matrix $R$ as defined in \eqref{eq:rat}, while for hybrid RS, we additionally provide the features of patients and doctors as separate matrices. Details on the calibration and hyperparameters of all methods are found in the Supplementary Material.

\paragraph{Evaluation}
We employ standard ranking metrics for the evaluation of RS and XML, namely Normalized Discounted Cumulative Gain@K (nDCG@K) \cite{article:HealthDoc_RSMF}, Precision@K (P@K) \cite{article:HealthDoc_RSContent}, and Recall@K \cite{article:HealthDoc_CF_Deng}. Additionally, we incorporate their propensity-scored counterparts of the first two (PSnDCG@K and PSP@K, respectively), which account for label popularity \cite{repo:XML}. That is, they prevent a good performance at the cost of predicting head labels and put emphasis on tail labels. In our setting, this is crucial as popular doctors are often already well-known to patients and may have longer waiting times. Thus, accurately recommending less popular doctors becomes highly relevant. Here, we represent PSnDCG@3 and Recall@10, but the remaining metrics lead to similar conclusions and can be found in the Supplementary Material.

\subsection{Results and Discussion}

\paragraph{Overall comparison: XML vs.\ RS} For methods with features (XML and LightFM) we select the most favorable scenario to compare the full capabilities of each method for both the seen and new patients (Figure~\ref{fig:Best_case_all}). SVD has poor performance in either case. BiVAE, without features, is competitive for seen patients but does not provide predictions for cold-case users. LightFM is the closest competitor to DECAF. Nonetheless, for seen patients, DECAF performs better by a considerable margin. For new patients, the difference in PS-nDCG is small, but DECAF presents a higher recall. A further comparison between LightFM and DECAF will be presented in next sections.

\paragraph{Comparing the Scenarios} We compare the performance of XML and LightFM (the remaining methods do not cope with features) across the different scenarios (Figure~\ref{fig:DECAFvslightfm}). For seen patients, it is clear that including more features is always beneficial regardless of the method, as S5 presents the highest scores. DECAF consistently outperforms LightFM, except for S1. Thus, LighFM copes better with a limited number of features as it takes advantage of its collaborative component. For new patients, the performance of LightFM does not show increased variation with the scenarios, while DECAF does. S4 is undoubtedly the most advantageous scenario for DECAF. This is easily explained as S2, S3 and S5 contain features that are not available for unseen patients and that were encoded with zero vectors. Comparing S4 with S1 tells us that the inclusion of distance features is beneficial. Thus, for seen users, DECAF can exploit existing features in the same or better way as LightFM, as long as there is a reasonable amount of them. For new users, LightFM takes limited advantage of existing features. DECAF is a better option if it is possible to appropriately encode the existing features such that they are present for new users, otherwise, LightFM is preferable.

\begin{figure}[h]
\centerline{\includegraphics[width=0.4\linewidth]{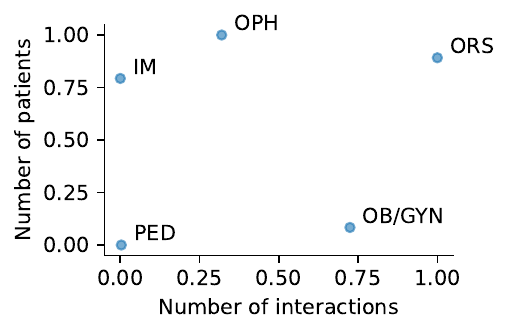}}
\caption{Profile of each specialization normalized with respect to the number of patients and number of interactions. }
\label{fig:SpecProfile}
\end{figure}

\begin{figure*}
\centerline{\includegraphics[width=0.8\linewidth]{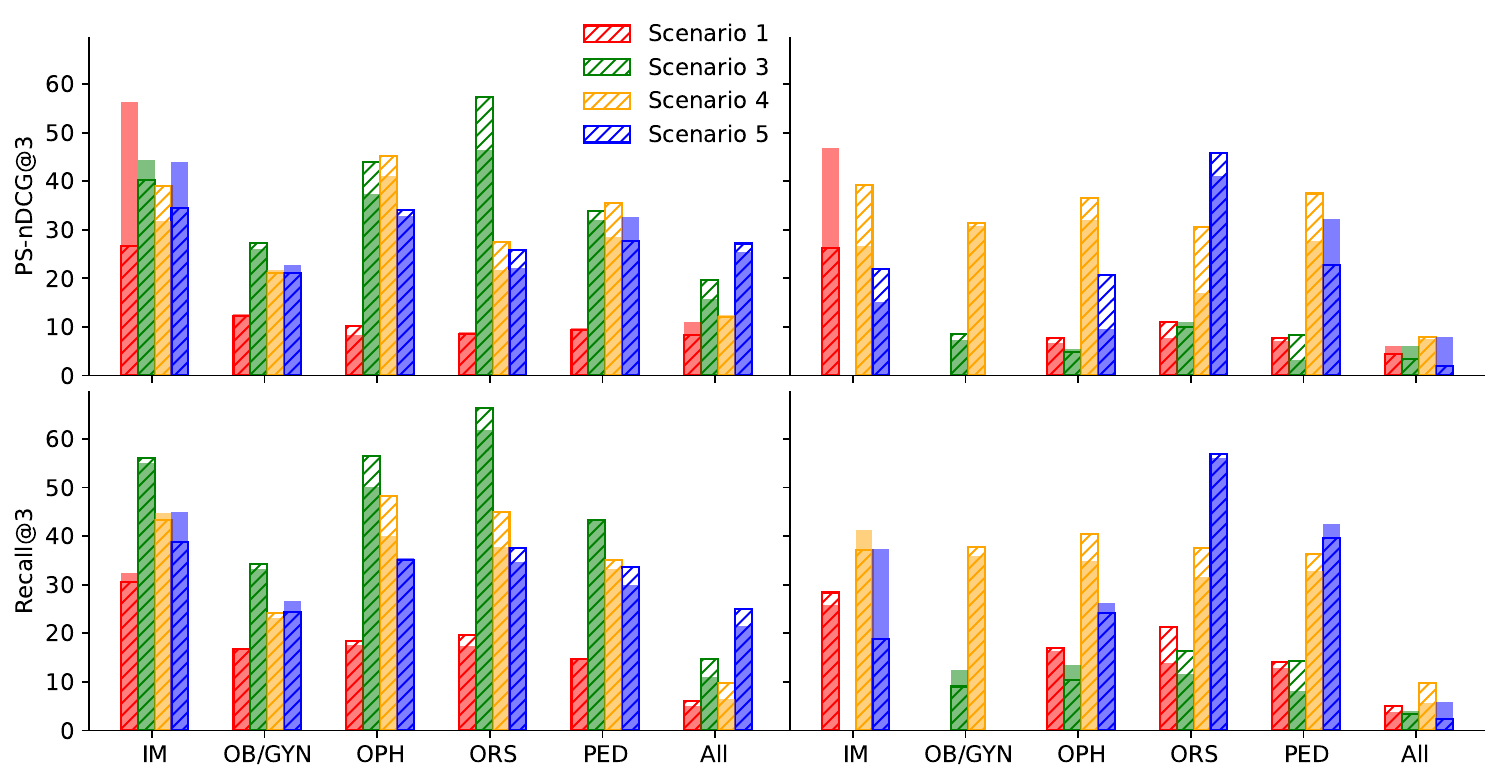}}
\caption{PS-nDCG@3 and Recall@3 for XML and LightFM for seen patients (left) and new patients (right). For each specialty (x-axis), we depict the score for each different scenario for lightFM in solid colors, and for XML in the same-color hatched bars. The absence of a bar indicates that XML does not provide at least 3 predictions for enough patients to have significant results.}
\label{fig:Scenarios_bar_overlayed}
\end{figure*} 

\paragraph{Medical Specializations.}

While the previous results evaluate the recommendations over the whole set of doctors, our goal is to predict within a predefined specialization. We select the specialization with the largest number of doctors, as they have a higher need for a recommendation tool. The five selected specializations have different profiles when it comes to the number of patients and the number of interactions (Figure~\ref{fig:SpecProfile}), which allows us to study the behavior of each method under different conditions. XML only provides the top-K labels for each training point (with $K=30$), while LightFM gives a complete ranked list of all the labels. To ensure a fair comparison, we compare the patients for whom we have at least 3 predictions per specialization in XML. Note that our goal is not to predict the correct specialization, but rather to recommend a suitable doctor for a given specialization (pre-determined by the family doctor).

By evaluating the results for different groups we can understand how their behavior differs from the pattern in the overall dataset and how this relates to the profiles of each specialization. We discuss in detail the comparison in Recall@10 and PS-nDCG@3 (Figure~\ref{fig:Scenarios_bar_overlayed}), but other metrics show similar results (see Supplementary Material). S2 does not provide enough predictions per specialization to allow for a significant comparison for unseen patients and it is not included in this section. This is explained by the lack of diversity in predictions, as most patients are labelled with doctors from the same specialization (see Supplementary Material for a more detailed analysis).

We will first consider \textbf{seen patients}. While S5 was the top performer for the overall dataset, for the considered categories S3 is now the most favourable setting. Although we did not select the groups based on interactions, specializations with the largest number of doctors necessarily imply more demand from patients and more frequent occurrence in the dataset. Therefore, S3 tends to shine in groups where previous historical information is more prevalent. This is further attested by the large difference in behaviour observed for ORS, the specialization with the highest number of interactions. Moreover, both S1 and S5 show similar results between the entire dataset and each group, while S3 and S4 present a considerable decrease. This tells us that regardless of the encoding, hospital information is mostly beneficial for the most popular specializations. Therefore, S5 remains the most suitable choice for patients with previous interactions, as it presents both a high performance and constant behaviour across specializations. Finally, we point out the behaviour of S1 for IM, in both seen and new patients. The distinctive characteristic of this group is that it includes a doctor with considerably more interactions, than any other in the same group (see Supplementary Material). The reduced amount of features in S1 is not enough for XML to overcome the popularity bias, while LightFM does so. However, with enough features (in any other scenario) this is not a limitation of XML.

For \textbf{new patients}, S4 remains the most favourable scenario, but the gap with respect to the others largely increased when compared to the overall prediction. This validates the choice of XML, as this is realistically the case in highest need for a recommendation tool, i.e., new patients requiring a consult in specializations with large number of doctors. The gap in performance between LightFM and DECAF holds for all groups except OB/GYN, where they show similar results. We note that this is the group with lower number of patients and highest interactions, so each patient has more interactions, where the collaborative component of LightFM shines. The need for two different models (S5 for seen patients and S5 for new patients) is evidently a limitation of this approach, but given the increase in performance it can be justifiable. Moreover, this study indicates that a more tuned selection of features can lead to an increased gap in performance.

% ####################################################################################################################

\section{Conclusion}
In this study, we tackled the complex problem of physician referral, in the absence of explicit feedback and with limited patient metadata. The inclusion of various specializations increased data redundancy and compounded the difficulty of the task. Importantly, our research also took into account the cold-start scenario for users, a critical aspect often neglected in analogous studies. To achieve this goal, we have recasted a traditional recommender setting into a multilabel classification problem that can be solved by current extreme classification methods. Also, we proposed a novel, unified model leveraging patient history across different specialties.

Unlike commonly used Recommender Systems (RS), our novel approach applies Extreme Multilabel Classification (XML). We demonstrated that XML takes advantage of available features, producing more pertinent top predictions compared to conventional RS. Nevertheless, the successful implementation of XML methods requires careful feature engineering to ensure that the features of patients and doctors occupy the same vector space. The study underscores the merits of XML over RS solutions across a variety of scenarios, each characterized by a different combination of four distinct feature groups. We propose strategic methods to map these groups into a shared vector space and illuminate the advantages and drawbacks of each combination of feature groups in relation to XML and RS solutions.

% ####################################################################################################################

\textbf{Funding}

This work was partially supported by FCT, IP, through project CMU/TIC/0016/2021

\bibliographystyle{unsrtnat}
\bibliography{references}  %%% Uncomment this line and comment out the ``thebibliography'' section below to use the external .bib file (using bibtex) .

%%% Uncomment this section and comment out the \bibliography{references} line above to use inline references.
% \begin{thebibliography}{1}

% 	\bibitem{kour2014real}
% 	George Kour and Raid Saabne.
% 	\newblock Real-time segmentation of on-line handwritten arabic script.
% 	\newblock In {\em Frontiers in Handwriting Recognition (ICFHR), 2014 14th
% 			International Conference on}, pages 417--422. IEEE, 2014.

% 	\bibitem{kour2014fast}
% 	George Kour and Raid Saabne.
% 	\newblock Fast classification of handwritten on-line arabic characters.
% 	\newblock In {\em Soft Computing and Pattern Recognition (SoCPaR), 2014 6th
% 			International Conference of}, pages 312--318. IEEE, 2014.

% 	\bibitem{hadash2018estimate}
% 	Guy Hadash, Einat Kermany, Boaz Carmeli, Ofer Lavi, George Kour, and Alon
% 	Jacovi.
% 	\newblock Estimate and replace: A novel approach to integrating deep neural
% 	networks with existing applications.
% 	\newblock {\em arXiv preprint arXiv:1804.09028}, 2018.

% \end{thebibliography}

\appendix

\section{Dataset details}

The dataset that forms the basis of this study is derived from patient-doctor consultations provided by a private European health network, with $P=1003809$ patients and $L=1054$ doctors. Each interaction involves a unique patient $p$ and doctor $l$, with assigned time and location. In addition, demographic data pertaining to both patients and doctors is available in separate files, along with the educational qualifications and specializations of the doctors. While there are initially over 50 specialities available in the data, many of them are too scarce and do not allow making meaningful insights from them, hence they need to be removed. In our experiments we keep only the specialties with more than $50$ doctors, which leaves us with a total of $5$ categories. These are listed by name in Table \ref{tab:medicine}, together with the corresponding abbreviations used throughout the paper. Additionally, we use the category ``\textit{All}", that includes all the specialities together --- note that in this last case the number of doctors of the single speciality is not the bottleneck, hence there we include also the doctors from fields other than  the five presented in Table \ref{tab:medicine}. 

\begin{figure}
\centering
\includegraphics[width=0.4\linewidth]{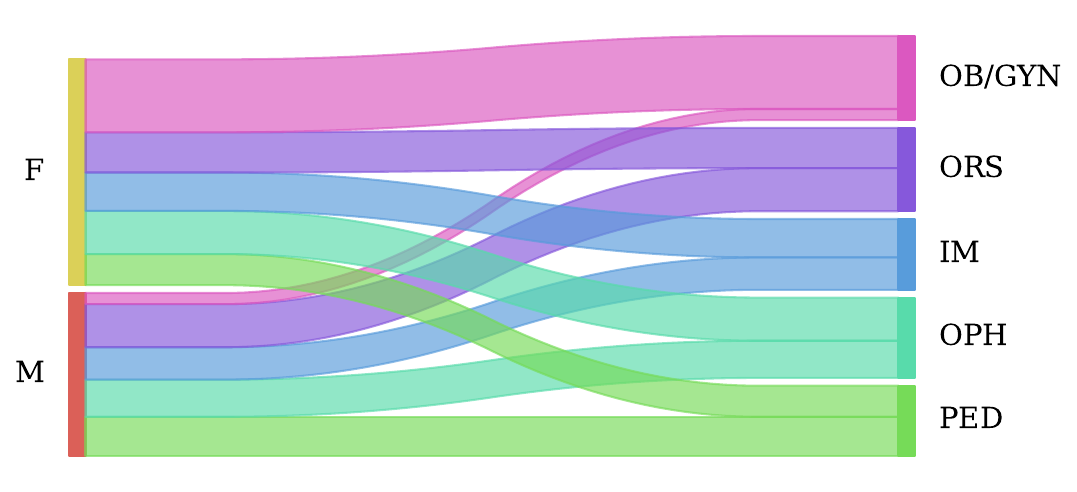}
\caption{The left side represents the gender distribution of patients (\textit{F - female, M - male}), and the right side indicates the specializations of the doctors they consulted.}
\label{fig:sex_spec}
\end{figure}

\begin{table}[b]
\caption{Medical Specialties and their Abbreviations}
\begin{center}
\begin{tabular}{r|l}
\hline
\textbf{Specialty} & \textbf{Abbreviation} \\
\hline
Internal Medicine & IM \\
Ophtalmology & OPH \\
Gynecology & OB/GYN \\
Pediatrics & PED \\
Orthopedics & ORS \\
\hline
\end{tabular}
\label{tab:medicine}
\end{center}
\end{table}
After this subsampling, we end up with a total number of $P=993458$ patients, $L=1044$ doctors and around $5.5$ million interactions.
Figure \ref{fig:sex_spec} shows the proportions of visits to each specialty based on the patients' sex. While the number of female patients is visibly larger than the number of male patients, the ratio of visits to different specialties are balanced, with the exception of the gynecology. 

\begin{figure}
\centerline{\includegraphics[width=0.4\linewidth]{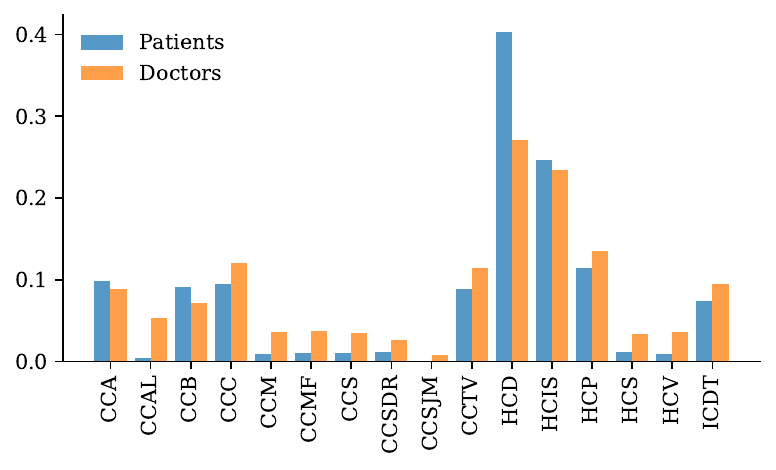}}
\caption{Fraction of patients/doctors visiting each hospital unit from the dataset. }
\label{fig:unidade_visits}
\end{figure}

Further, the dataset encompasses $16$ different hospitals with attributed geographical locations. The same doctor might be operating across multiple locations, and patients may attend different hospitals depending on the nature of the appointment or the distance to the hospital. Figure~\ref{fig:unidade_visits} illustrates the variations in popularity among hospitals. While the popularity of units differs significantly, the ratio between doctors and patients in each unit is often similar, as might be expected.

\subsection{Trainig-Test Split}

The interactions data is split with respect to patients, and taking into account the times of the visits. In particular, we take a single patient $p$, and choose a point in time $t$ as a threshold. Then all the interactions of that patient that happened before $t$ are sent to the training set, while all that come after are sent to the test set. This is done to avoid ``\textit{peaking into the future}", i.e., having the data about the patient that come chronologically after the interaction we want to predict. It is additionally important to note that a certain number of patients in the test set was never seen in the training, which allows to evaluate the methods under the cold-start problem. Hence, the test set is split into two, one for the patients that were already seen in the training (\textit{seen patients}), and one for the others (\textit{new patients}), and these two are evaluated separately throughout the paper. The exact sizes of the partitions are given in Table 2 of the main paper.

\subsection{Feature Normalization} We derive four distinct feature groups from the interactions data, as explained in Section 4 (Feature Extraction) of the main paper. Here we specify how the normalization of the specific features was performed. In the first group of features, $\textbf{b}^p, \textbf{b}^d$, the second coordinates $b^p_2, b^d_2 \in \mathbb{N}$ denote age of the subject; these are normalized dividing by the maximum age present in the training set. On the other side, the third coordinates, $b^p_3, b^d_3 \in \mathbb{N}$, that give the number of occurrences of the patient/doctor in the training set, will have significantly different ranges for the two groups; hence, we normalize them separately. $b^p_3$ is divided by the maximum value over all the patients, and $b^d_3$ by the maximum over all doctors.

\subsection{Doctor Specialities}
In order to evaluate specific specialties, we split the label matrix $Y$ introduced in eq. (2) of the main paper accordingly. Given $Y$ and the binary matrix of predicted labels $Y_{\textrm{pred}}$, for specialty $s$, we take the submatrices of $Y$ and  $Y_{\textrm{pred}}$ that refer to doctors of specialty $s$. We also remove patients with no positive labels for specialty $s$ giving the subsampled matrices $Y_{\textrm{pred}}^s$ and $Y^s$ for each specialty. 

\section{Details of the XML model used}

The XML algorithm used in our experiments is DECAF: Deep Extreme Classification with Label Features \cite{article:XML_DECAF}, and a detailed description of the algorithm can be found within the original paper, and a specific Python implementation used is documented in their github page \texttt{github.com/Extreme-classification/DECAF}. We selected the \textit{DECAF-lite} version of the method, due to computational limitations. Regarding hyperparameters, we have kept those suggested by the authors for the smallest dataset \textit{LF-Amazon-131K}, with a few exceptions listed in Table~\ref{table:parametersDECAF}. These hyperparameters were only modified when they were strictly related to the number of labels or the number of features, which differs between our dataset and \textit{LF-Amazon-131K}. We have altered them in the same proportion as the difference in number of labels and they were not further calibrated. Due to the difference in number of samples and for computational reasons we also reduced the batch size.

\begin{table}[ht]
\centering
\caption{Values of the hyperparameters for DECAF.}
\begin{tabular}{@{}ll@{}}
\toprule
Description & Hyperparameter Value \\
\midrule
Embedding dimension  & \( \texttt{embadding\_dims} = 32 \) \\
Factor for number of clusters        & \( \texttt{b\_factors} = 7 \) \\
Beam size & \( \texttt{beam\_size} = 30 \) \\
Batch size & \( \texttt{batch\_sizes} = 20 \) \\
\bottomrule
\end{tabular}
\label{table:parametersDECAF}
\end{table}

\section{Calibration of the benchmarks}

The Python implementations of the Recommender System benchmarks used in our experiments are built by Microsoft \cite{MicrosoftRecommenders}, and detailed documentations are available on their github page \texttt{github.com/microsoft/recommenders}. In this section we give list of hyperparameter values used for each of the benchmarks, as well as for the XML method. In particular, the hyperparameters used for XML are given in Table \ref{table:parametersDECAF}; hyperparameters for SVD are in Table \ref{table:parametersSVD}; hyperparameters for BiVAE are in Table \ref{table:parametersBiVae}; hyperparameters for LightFM are in Table \ref{table:parametersLFM}.

\begin{table}[ht]
\centering
\caption{Values of the hyperparameters for training SVD.}
\begin{tabular}{@{}ll@{}}
\toprule
Description & Hyperparameter Value \\
\midrule
Learning rate        & \( \texttt{lr\_all} = 0.004 \) \\
No of latent factors & \( \texttt{n\_factors} = 32 \) \\
No of epochs         & \( \texttt{n\_epochs} = 40 \) \\
\bottomrule
\end{tabular}
\label{table:parametersSVD}
\end{table}

\begin{table}[ht]
\centering
\caption{Values of the hyperparameters for training BiVae.}
\begin{tabular}{@{}ll@{}}
\toprule
Description & Hyperparameter Value \\
\midrule
No of latent factors  & \( \texttt{k} = 32 \) \\
No of epochs          & \( \texttt{n\_epochs} = 40 \) \\
Dimension of encoder  & \( \texttt{encoder\_structure} = 32 \) \\
\bottomrule
\end{tabular}
\label{table:parametersBiVae}
\end{table}

\begin{table}[ht]
\centering
\caption{Values of the hyperparameters for training LightFM.}
\begin{tabular}{@{}ll@{}}
\toprule
Description & Hyperparameter Value \\
\midrule
No of latent factors         & \( \texttt{no\_components} = 32 \) \\
Learning rate         & \( \texttt{learning\_rate} = 0.01 \) \\
Loss         & \( \texttt{loss} = \texttt{'warp'} \) \\
\bottomrule
\end{tabular}
\label{table:parametersLFM}
\end{table}

The hyperparameters not specifically mentioned in the above tables are left to their default value. The same number of latent factors is used for all the methods. We note here that we did not engage into an extensive cross-validation for finding the best possible set of parameters for each method, since the main focus of our research was to show how XML can be applied in the recommendation task beyond language processing.

\subsection{Computational setting}

The experiments were executed on a user level 64-bit machine with the  Intel Broadwell CPU, NVIDIA T4 GPU, 200GB SSD, and 12 GB of RAM. 

\section{Additional Results}

\subsection{Additional Metrics} 

In the main paper we show the metric results of the propensity-scored normalized discounted cumulative gain PSnDCG@3, and Recall@3 and Recall@10, as the principled metrics of success in the related literature. Additionally, here we include nDCG@3, as well as the Precision P@3 and propensity-scored precision PS-P@3 (Tables \ref{tab:seen} and \ref{tab:unseen}). All the metrics indicate that DECAF is superior to other methods, and in specific, it works the best under the Scenario 5 for the seen patients, and under the Scenario 4 for the unseen. 

\begin{table*} \caption{Results for patients seen in train for XML and LightFM. The maximum values of each column are highlighted in bold.}\label{tab:seen} \centering
\scriptsize
\begin{tabular}{llrrrrrr}
\toprule
    &    &  PS-P@3 &  PS-nDCG@3 &   P@3 &  nDCG@3 &  Recall@3 &  Recall@10 \\
\midrule
DECAF & S1 &   10.29 &       8.31 & 12.51 &   13.31 &      6.04 &      14.00 \\
& S2 &   20.72 &      19.40 & 23.47 &   27.56 &       NaN &        NaN \\
    & S3 &   21.68 &      19.59 & 22.17 &   24.80 &     14.76 &      32.44 \\
    & S4 &   13.94 &      12.00 & 16.93 &   18.47 &      9.79 &      23.28 \\
    & S5 &   \textbf{28.14} &     \textbf{ 27.18} & \textbf{30.23} &  \textbf{ 36.47} &     \textbf{24.98} &      \textbf{46.10} \\
LightFM & S1 &   11.58 &      10.98 & 10.45 &   10.97 &      4.98 &      11.57 \\
 & S2 &    3.86 &       3.28 &  3.88 &    4.05 &       NaN &        NaN \\
    & S3 &   16.97 &      15.69 & 16.74 &   18.43 &     10.90 &      25.92 \\
    & S4 &   13.18 &      12.35 & 12.29 &   13.04 &      6.46 &      16.96 \\
    & S5 &   25.75 &      25.36 & 26.24 &   31.34 &     21.42 &      41.96 \\
BiVAE & - &   18.46 &      17.46 & 22.60 &   26.82 &     14.72 &      26.42 \\
SVD & - &    0.19 &       0.15 &  0.20 &    0.20 &      0.16 &       0.70 \\
\bottomrule
\end{tabular}

\end{table*}
% UNSEEN
\begin{table*}\caption{Results for new patients for XML and LightFM. The maximum values of each column are highlighted in bold.}\label{tab:unseen} \centering
\scriptsize
\begin{tabular}{llrrrrrr}
\toprule
    &    &  PS-P@3 &  PS-nDCG@3 &  P@3 &  nDCG@3 &  Recall@3 &  Recall@10 \\
\midrule
DECAF & S1 &    6.50 &       4.42 & 5.36 &    6.70 &      5.09 &      12.76 \\
    & S2 &    3.37 &       2.40 & 3.34 &    4.43 &       NaN &        NaN \\
    & S3 &    4.49 &       3.29 & 3.38 &    4.43 &      3.41 &       7.97 \\
    & S4 &   \textbf{10.64} &       7.82 & \textbf{7.81} &   \textbf{10.89} &      \textbf{9.78 }&     \textbf{ 23.10} \\
    & S5 &    2.51 &       1.95 & 1.73 &    2.48 &      2.32 &       5.89 \\
LightFM & S1 &    7.95 &       6.06 & 4.22 &    5.20 &      3.69 &       9.22 \\
& S2 &    1.90 &       1.32 & 1.19 &    1.48 &       NaN &        NaN \\
    & S3 &    8.20 &       6.13 & 4.43 &    5.40 &      4.08 &       8.59 \\
    & S4 &    9.79 &       7.49 & 5.41 &    6.90 &      5.59 &      15.85 \\
    & S5 &    9.98 &       \textbf{7.85} & 5.54 &    7.35 &      5.88 &      16.14 \\
BiVAE & - &   - &       - & - &    - &      - &       - \\
SVD & - &    0.61 &       0.42 & 0.45 &    0.57 &      0.42 &       1.10 \\
\bottomrule
\end{tabular}
\end{table*}

\begin{figure}
\centerline{\includegraphics[width=0.6\linewidth]{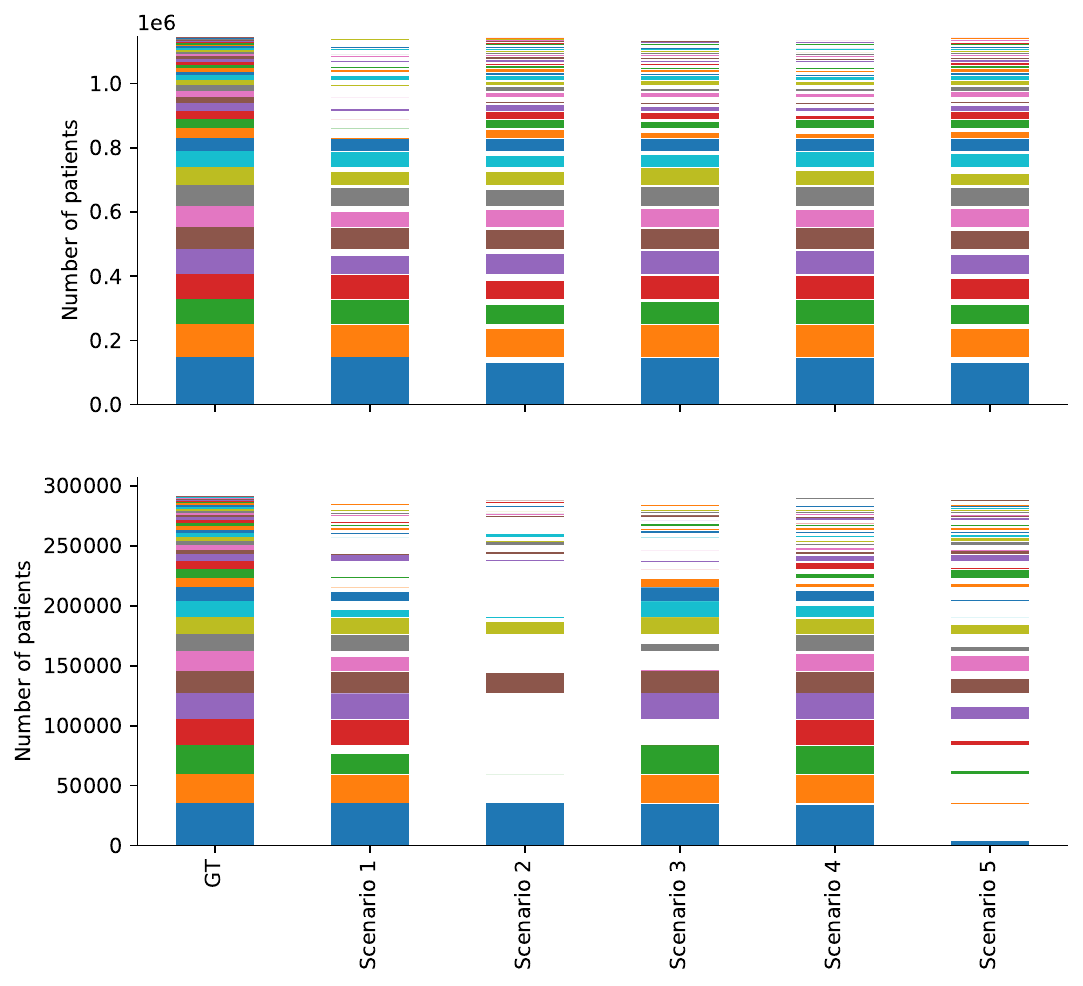}}
\caption{Number of patients with at least one label of each specialization in the test, for seen (top) and unseen (bottom) patients. For each Scenario, we depict all of the specializations stacked with different colors.}
\label{fig:PredictionsPerSpec}
\end{figure}

\subsection{Further insights on XML}

Keeping in mind that XML only provides a limited number of relevant labels, we are interested in understanding how they are distributed between the different doctors for each scenario.  While the previous results show how accurate the top results are, here we want to see how diverse the 30 predictions returned by DECAF are. This is relevant as we wish to predict different specializations, meaning that some specializations might not even be present in the 30 predictions given by DECAF. In Figure \ref{fig:PredictionsPerSpec}, we see that predictions for the new patients, in Scenarios 2 and 5, do not contain all of the specialties for the large percentage of patients.

\begin{figure*}
\centerline{\includegraphics[width=\linewidth]{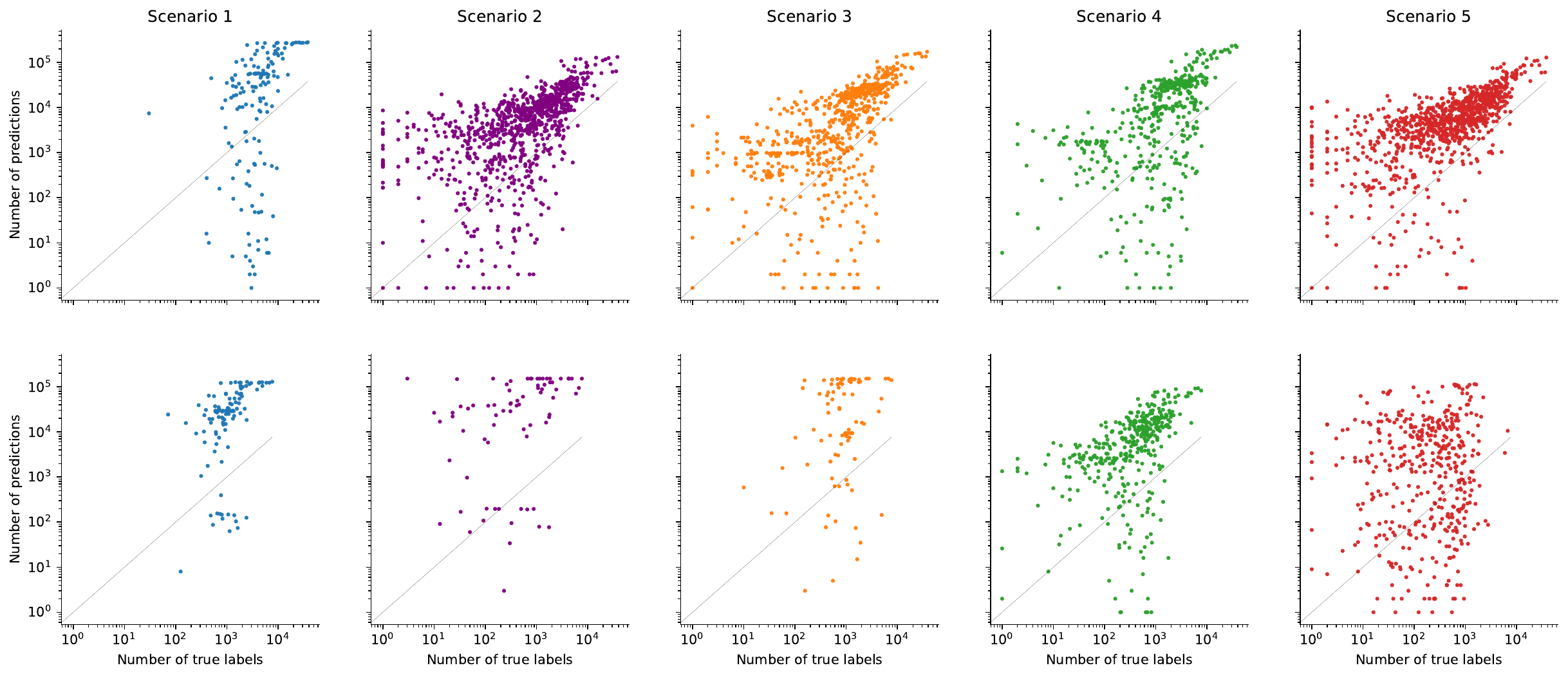}}
\caption{Predictions of DECAF across scenarios. Each point corresponds to a different doctor, where the x-coordinate is the number of times this doctor appears as a label in the test dataset, and the y-coordinate is the number of times DECAF predicts them. The results of the seen patients are shown on top, and the new patients are on bottom.}
\label{fig:PredictionsPerDoc}
\end{figure*}

To understand this better, we look in Figure \ref{fig:PredictionsPerDoc}.
For seen patients, all scenarios except for S1 (Scenario 1) reflect the real distribution of the ground truth, with most doctors predicted with a frequency similar to that shown in the test set. The limited features in S1 lead to the prediction of the most popular doctors only. For new patients, only S4 can maintain this pattern. In this case, S1, S2, and S3 predict few doctors with a large frequency, so there is small variability in the recommendations. However, it is noticeable that S1 suffers the least, as all features are present during test time. S5 shows a different behavior, where the most popular doctors are not predicted. Similarly to S2 and S3, new users in S5 have a large component of features set to zero. However, the addition of distance information leads to the prediction of less popular doctors. Given the high performance of S2 for seen patients, we hypothesise that in S5 classifiers for the most popular doctors rely mostly on specialization features, while those for tail doctors rely on distance information.

\end{document}